# $f$-Divergence constrained policy improvement

**Boris Belousov**                                                                 BELOUSOV@IAS.TU-DARMSTADT.DE
**Jan Peters**                                                                     PETERS@IAS.TU-DARMSTADT.DE
*Department of Computer Science*
*Technische Universität Darmstadt*
*FG IAS, Hochschulstr. 10, 64289 Darmstadt, Germany*

## Abstract

To ensure stability of learning, state-of-the-art generalized policy iteration algorithms augment the policy improvement step with a trust region constraint bounding the information loss. The size of the trust region is commonly determined by the Kullback-Leibler (KL) divergence, which not only captures the notion of distance well but also yields closed-form solutions. In this paper, we consider a more general class of $f$-divergences and derive the corresponding policy update rules. The generic solution is expressed through the derivative of the convex conjugate function to $f$ and includes the KL solution as a special case. Within the class of $f$-divergences, we further focus on a one-parameter family of $\alpha$-divergences to study effects of the choice of divergence on policy improvement. Previously known as well as new policy updates emerge for different values of $\alpha$. We show that every type of policy update comes with a compatible policy evaluation resulting from the chosen $f$-divergence. Interestingly, the mean-squared Bellman error minimization is closely related to policy evaluation with the Pearson $\chi^2$-divergence penalty, while the KL divergence results in the soft-max policy update and a log-sum-exp critic. We carry out asymptotic analysis of the solutions for different values of $\alpha$ and demonstrate the effects of using different divergence functions on a multi-armed bandit problem and on common standard reinforcement learning problems.

## 1. Introduction

Many state-of-the art reinforcement learning algorithms, including natural policy gradients (Kakade, 2001; Bagnell and Schneider, 2003; Peters et al., 2003), trust region policy optimization (TRPO) (Schulman et al., 2015) and relative entropy policy search (REPS) (Peters et al., 2010), impose a Kullback-Leibler (KL) divergence constraint between successive policies during parametric policy iteration to avoid large steps towards unknown regions of the state space. Similar objective functions with an entropy-like term have been proposed in the context of linearly-solvable optimal control (Kappen, 2005; Todorov, 2006) and inverse reinforcement learning (Ziebart et al., 2008). In all these approaches, the objective function has the form of the free energy, therefore they all can be viewed as performing free energy minimization (Still and Precup, 2012).

In this paper, we explore the implications of using a generic $f$-divergence to constrain the policy improvement. The objective function in this case resembles the free energy objective, commonly encountered in variational methods (Wainwright and Jordan, 2007), but with the KL divergence replaced by an $f$-divergence. The idea of using $f$-divergence penalties for general non-linear problems goes back to the early work by Teboulle (1992). Generalizations of the KL divergence—apart from being very useful at providing a unified view of existing algorithms, e.g., statistical inference can be seen as $f$-divergence minimization (Altun and Smola, 2006) and various message passing algorithms can be understood as minimizing different $\alpha$-divergences (Minka, 2005),—pave the way to new algorithms that are better suited for particular problems—e.g., the generative adversarial networks (GANs) (Goodfellow et al., 2014) minimize a particular $f$-divergence, the Jensen-Shannon divergence, which is well-suited for measuring similarity between real-world images (Nowozin et al., 2016). This paper is, to our knowledge, the first to shed light on policy improvement in reinforcement learning from the $f$-divergence point of view.

## 2. Background

This section provides the background on the $f$-divergence, the $\alpha$-divergence, and the convex conjugate function, highlighting the key properties required for our derivations.

The **$f$-divergence** (Csiszár, 1963; Morimoto, 1963; Ali and Silvey, 1966) generalizes many similarity measures between probability distributions (Sason and Verdu, 2016). For two distributions $\pi$ and $q$ on a finite set $\mathcal{A}$, the $f$-divergence is defined as

$$D_f(\pi\|q) \triangleq \sum_{a\in\mathcal{A}} q(a) f\left(\frac{\pi(a)}{q(a)}\right),$$

where $f$ is a convex function on $(0,\infty)$ such that $f(1) = 0$. For example, the KL divergence corresponds to $f_{KL}(x) = x\log x$. Note that $\pi$ must be absolutely continuous with respect to $q$ to avoid division by zero, i.e., $q(a) = 0$ implies $\pi(a) = 0$ for all $a \in \mathcal{A}$. We additionally assume $f$ to be continuously differentiable, which includes all cases of interest for us. The $f$-divergence can be generalized to *unnormalized distributions*. For example, the generalized KL divergence (Zhu and Rohwer, 1995) corresponds to $f_1(x) = x\log x - (x-1)$. The derivations in this paper benefit from employing unnormalized distributions and subsequently imposing the normalization condition as a constraint.

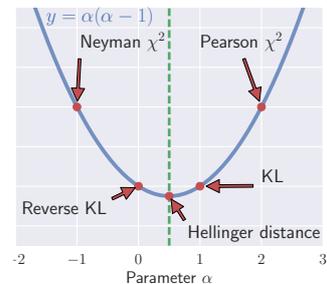

Figure 1: The $\alpha$-divergence smoothly connects several prominent divergences.

The **$\alpha$-divergence** (Chernoff, 1952; Amari, 1985) is a one-parameter family of $f$-divergences generated by the $\alpha$-function $f_\alpha(x)$ with $\alpha \in \mathbb{R}$. The particular choice of the family of functions $f_\alpha$ is motivated by generalization of the natural logarithm (Cichocki and Amari, 2010). The $\alpha$-logarithm $\log_\alpha(x) \triangleq (x^{\alpha-1} - 1)/(\alpha - 1)$ is a power function for $\alpha \neq 1$ that turns into the natural logarithm for $\alpha \to 1$. Replacing the natural logarithm in the derivative of the KL divergence $f_1' = \log x$ by the $\alpha$-logarithm and integrating $f_\alpha'$ under the condition that $f_\alpha(1) = 0$ yields the $\alpha$-function

$$f_\alpha(x) \triangleq \frac{(x^\alpha - 1) - \alpha(x - 1)}{\alpha(\alpha - 1)}. \tag{1}$$

| Divergence | $\alpha$ | $f(x)$ | $f'(x)$ | $(f^*)'(y)$ | $f^*(y)$ | dom $f^*$ |
|---|---|---|---|---|---|---|
| KL | $1$ | $x \log x - (x-1)$ | $\log x$ | $e^y$ | $e^y - 1$ | $\mathbb{R}$ |
| Reverse KL | $0$ | $-\log x + (x-1)$ | $-\frac{1}{x} + 1$ | $\frac{1}{1-y}$ | $-\log(1-y)$ | $y < 1$ |
| Pearson $\chi^2$ | $2$ | $\frac{1}{2}(x-1)^2$ | $x - 1$ | $y + 1$ | $\frac{1}{2}(y+1)^2 - \frac{1}{2}$ | $y > -1$ |
| Neyman $\chi^2$ | $-1$ | $\frac{(x-1)^2}{2x}$ | $-\frac{1}{2x^2} + \frac{1}{2}$ | $\frac{1}{\sqrt{1-2y}}$ | $-\sqrt{1-2y} + 1$ | $y < \frac{1}{2}$ |
| Hellinger | $\frac{1}{2}$ | $2(\sqrt{x} - 1)^2$ | $2 - \frac{2}{\sqrt{x}}$ | $\frac{4}{(2-y)^2}$ | $\frac{2y}{2-y}$ | $y < 2$ |

Table 1: Function $f_\alpha$, its convex conjugate $f_\alpha^*$, and their derivatives for some values of $\alpha$

The $\alpha$-divergence generalizes the KL divergence, reverse KL divergence, Hellinger distance, Pearson $\chi^2$-divergence, and Neyman (reverse Pearson) $\chi^2$-divergence. Figure 1 displays well-known $\alpha$-divergences as points on the parabola $y = \alpha(\alpha - 1)$. For every divergence, there is a reverse divergence symmetric with respect to the point $\alpha = 0.5$, corresponding to the Hellinger distance.

The **convex conjugate** of $f(x)$ is defined as $f^*(y) = \sup_{x \in \mathrm{dom}\, f}\{\langle y, x\rangle - f(x)\}$, where the angle brackets $\langle y, x\rangle$ denote the dot product (Boyd and Vandenberghe, 2004). The key property $(f^*)' = (f')^{-1}$ relating the derivatives of $f^*$ and $f$ yields Table 1, which lists common functions $f_\alpha$ together with their convex conjugates and derivatives. In the general case (1), the convex conjugate and its derivative are given by

$$f_\alpha^*(y) = \frac{1}{\alpha}(1 + (\alpha-1)y)^{\frac{\alpha}{\alpha-1}} - \frac{1}{\alpha},$$
$$(f_\alpha^*)'(y) = \sqrt[\alpha-1]{1 + (\alpha-1)y},$$
$$\text{for } y(1-\alpha) < 1. \tag{2}$$

Function $f_\alpha$ is convex, non-negative, and attains minimum at $x = 1$ with $f_\alpha(1) = 0$. Function $(f_\alpha^*)'$ is positive on its domain with $(f_\alpha^*)'(0) = 1$. Function $f_\alpha^*$ has the property $f_\alpha^*(0) = 0$. The linear inequality constraint (2) on the dom $f_\alpha^*$ follows from the requirement dom $f_\alpha = (0, \infty)$. Another result from convex analysis crucial to our derivations is *Fenchel's equality*

$$f^*(y) + f(x^\star(y)) = \langle y, x^\star(y)\rangle, \tag{3}$$

where $x^\star(y) = \arg\sup_{x \in \mathrm{dom}\, f}\{\langle y, x\rangle - f(x)\}$. We will occasionally put the conjugation symbol at the bottom, especially for the derivative of the conjugate function $f'_* \triangleq (f^*)'$.

## 3. Policy improvement with $f$-divergence for multi-armed bandits

In order to develop intuition regarding the influence of entropic penalties on policy improvement, we first consider a simplified version of the reinforcement learning problem—namely, the stochastic multi-armed bandit problem (Bubeck and Cesa-Bianchi, 2012). The resulting algorithm is closely related to the family of Exp3 algorithms (Auer et al., 2003), which are extremely good on adversarial bandit problems but have fallen behind the state-of-the-art for stochastic bandit problems. We nevertheless focus on the stochastic bandit problems to illustrate crucial effects of entropic penalties on policy improvement in a clean intuitive way. This section is meant to serve as a foundation for the reinforcement learning section.

At every time step $t \in \{1, \ldots, T\}$, an agent chooses among $K$ actions $a \in \mathcal{A}$ and after every choice $a_t = a$ receives a noisy reward $r_t = r(a_t, \epsilon_t)$ drawn from a distribution with mean $Q(a) = \mathbb{E}_\epsilon[r(a, \epsilon)]$. The goal of the agent is to maximize the expected total reward $J \triangleq \mathbb{E}[\sum_{t=1}^T r_t]$. If the agent knew the true value of each action $Q(a)$, the optimal strategy would be to always choose the action with the highest value, $a_t^* = \arg\max_a Q(a)$. However, since the agent does not know $Q$ in advance, it faces the exploration-exploitation dilemma at every time step. It can either pick the best action according to the current estimate of action values $\tilde{Q}_t$ or try an exploratory action that may turn out to be better than $\arg\max \tilde{Q}_t$. A generic way to encode the exploration-exploitation trade-off is by introducing a policy $\pi_t$, that is, a distribution over $a \in \mathcal{A}$ from which the agent draws an action $a_t \sim \pi_t$. Thus, the question becomes: given the current policy $\pi_t$ and the current estimate of action values $\tilde{Q}_t$, what should the policy $\pi_{t+1}$ at the next time step be? Unlike the choice of the best action under perfect information, such sampling policies are hard to derive from first principles (Ghavamzadeh et al., 2015).

In this section, we formalize the exploration-exploitation trade-off by choosing the next stochastic policy to maximize the expectation of $\tilde{Q}_t$ while regularizing the change in the policy through the $f$-divergence. This formalization always yields both a convex optimization problem and, as its solution, the next stochastic policy. For the $\alpha$-divergence, we obtain both classical policy updates (e.g., $\epsilon$-greedy, softmax, etc.) as well as novel ones resulting from different $\alpha$'s. Empirical evaluations on a simulated multi-armed bandit environment illustrate the differences in policy updates corresponding to various penalty functions.

### 3.1 $f$-Divergence constrained policy optimization problem for bandits

A policy improvement at time step $t$ is derived by maximizing the expected return $J(\pi) \triangleq \mathbb{E}_{a \sim \pi}[Q(a)]$ while punishing the policy change between the old policy $q(a) \triangleq \pi_t(a)$ and the new policy $\pi(a) \triangleq \pi_{t+1}(a)$ by an $f$-divergence as a kind of regularization. This regularization is added to the objective function weighted by a 'temperature' $\eta$ to trade-off exploration vs exploitation. For simplicity, the current estimate of action values is denoted by $Q(a) \triangleq \tilde{Q}_t(a)$. A policy $\pi$ that achieves a higher expected return than $q$ is a solution of the following optimization problem

$$\begin{aligned} \max_\pi \quad & J_\eta(\pi) = \sum_a Q(a)\pi(a) - \eta \sum_a q(a) f\left(\frac{\pi(a)}{q(a)}\right), \\ \text{s.t.} \quad & \sum_a \pi(a) = 1, \\ & \pi(a) \geq 0, \ \forall a \in \mathcal{A}. \end{aligned} \quad (4)$$

Different formulations of Problem (4) with the KL divergence in place of the $f$-divergence have been proposed. An inequality constraint (Peters et al., 2010) bounding the information loss by certain fixed amount $\epsilon$ can automatically yield an $\eta$ as a Lagrange multiplier found during optimization. Alternating between maximization of the expected return at fixed information loss and subsequent information loss minimization at constant expected return (Still and Precup, 2012) yields algorithms that allow for automatic adaptation of $\epsilon$. In (4), the divergence is added as a penalty, as suggested by Azar and Kappen (2012). Since the treatment of these three mathematical problems differs only in the sets of hyperparameters, our results straightforwardly generalize to the other two formulations.

Differentiating the Lagrangian of problem (4)

$$L_\eta(\pi, \lambda, \kappa) = \sum_a Q(a)\pi(a) - \eta \sum_a q(a) f\left(\pi(a)/q(a)\right) - \lambda \left(\sum_a \pi(a) - 1\right) + \sum_a \kappa(a)\pi(a)$$

with respect to $\pi(a)$ and equating the derivative to zero, we find an expression for the optimal new policy $\pi^\star$ through the derivative of the convex conjugate function

$$\pi^\star(a) = q(a) f'_*\left(\frac{Q(a) - \lambda + \kappa(a)}{\eta}\right). \tag{5}$$

Substituting $\pi^\star$ back into the Lagrangian and using Fenchel's equality (3), we arrive at the dual problem

$$\begin{aligned}
\min_{\lambda, \kappa} \quad & g(\lambda, \kappa) = \eta \sum_a q(a) f^*\left(\frac{Q(a) - \lambda + \kappa(a)}{\eta}\right) + \lambda, \\
\text{s.t.} \quad & \kappa(a) \geq 0, \quad \forall a \in \mathcal{A}, \\
& \arg f^* \in \text{range}_{x \geq 0} f'(x), \quad \forall a \in \mathcal{A}.
\end{aligned} \tag{6}$$

The constraint on the argument of the convex conjugate has a particularly simple linear form (2) for the $\alpha$-function $f_\alpha(y)$; see Section 2 for the necessary background. In the very special KL divergence case it completely disappears because range $f'_1 = \mathbb{R}$, see Table 1 for well-known divergences. In the multi-armed bandit scenario, one is free to choose whether to solve the optimization problem in the primal form (4) or in the dual form (6). However, as we will see later, the dual formulation (6) is much better suited for reinforcement learning.

**When the Lagrange multipliers $\kappa$ disappear.** When choosing function $f$, it is important to consider whether $0 \in \text{dom } f'(x)$, as it implies that the improved policy can put zero mass on some actions. Although $f$ is not required to have a finite value at $x = 0$ by definition, for many divergences it does (e.g., for any $\alpha$-divergence for $\alpha > 1$) or it has a finite limit (e.g., for the KL divergence). However, even those functions $f$ that are defined at zero do not necessarily have a finite derivative at zero (e.g., for the KL-divergence $f'_1(x) = \log x$). If $0 \notin \text{dom } f'(x)$, the new policy $\pi$ cannot put zero probability mass on actions that were possible under the old policy $q$. In other words, if $q(a) > 0$, then $\pi(a) > 0$, and the complementary slackness condition $\kappa(a)\pi(a) = 0$ implies $\kappa(a) = 0$. Therefore, *we can ignore the non-negativity constraint if $f'$ has a singularity at $x = 0$*. Thereby, one commonly omits the non-negativity constraint when using the KL divergence or the reverse KL. In general, $\kappa(a) = 0$ for any $f_\alpha$ with $\alpha \leq 1$. On the other hand, if $0 \in \text{dom } f'(x)$, some actions may get zero probability under the new policy, and $\kappa(a) \neq 0$. This case emerges for $f_\alpha$ with $\alpha > 1$.

### 3.2 Effects of the exploration-exploitation-tradeoff and the divergence type

The exploration-exploitation parameter $\eta$ controls the step size, while the divergence type $\alpha$ affects the search direction.

**Greedy policy vs no change: effects of the temperature parameter $\eta$.** Independent of the $f$-divergence chosen, the temperature parameter $\eta$ controls the magnitude of deviation of the new policy from the old one. The optimal Lagrange multiplier $\lambda^\star$ plays the role of the baseline similar as in policy gradient methods (Sutton et al., 2000; Peters and Schaal, 2006). For low temperatures $\eta \to 0$, the policy $\pi^\star(a) \to \delta(a, \arg\max_a Q(a))$ tends towards greedy action selection (Sutton and Barto, 1998) with $\lambda^\star \to \max_a Q(a)$. For $\eta \to \infty$, both the policy $\pi^\star(a) \to q(a)$ and the baseline $\lambda^\star \to \sum_a q(a)Q(a)$ do no longer change. For other values of $\eta$, the agent chooses an intermediate solution between focusing on a single best action and remaining close to the old policy.

**From $\epsilon$-greedy to $\epsilon$-elimination: effects of $\alpha$ on policy improvement.** Given the functional form of the policy update (5), we can find the Lagrange multipliers $\lambda$ and $\kappa$ and thus obtain a closed-form solution of Problem (4) in two cases: when $f'_*$ is exponential or linear, which corresponds to $\alpha = 1$ and $\alpha = 2$. For other $\alpha$'s, we have to solve the optimization problem numerically. Table 2 summarizes the effects of $\alpha$ on policy improvement. For $\alpha \ll 0$, we recover the $\epsilon$-**greedy** action selection method (Sutton and Barto, 1998), which assigns probability $1-\epsilon$ to the best action and spreads the remaining probability mass uniformly among the rest; parameter $\eta$ is proportional to $\epsilon$, see Section 3.2. Values of $\alpha \leq 1$ correspond to $\epsilon$-**soft** policies (Sutton and Barto, 1998); parameter $\alpha$ controls the distribution of probability mass among suboptimal actions: large negative values of $\alpha$ correspond to more uniform distributions, whereas smaller $\alpha$'s distribute the probability mass according to values. When $\alpha = 1$, we arrive at a form of the **soft-max** policy (Bridle, 1990) with the log-sum-exp baseline

Table 2: Effects of $\alpha$ on the policy $\forall \eta > 0$

| Policy $\pi$ | $\alpha$ |
|---|---|
| greedy | $-\infty$ |
| $\epsilon$-greedy | $\alpha \ll 0$ |
| $\epsilon$-soft | $\alpha \leq 1$ |
| soft-max | $\alpha = 1$ |
| linear | $\alpha = 2$ |
| $\epsilon$-elimin. | $\alpha \gg 1$ |
| elimination | $+\infty$ |

$$\pi^\star(a) = q(a)e^{\frac{1}{\eta}(Q(a)-\lambda^\star)} = \frac{q(a)e^{\frac{1}{\eta}Q(a)}}{\sum_b q(b)e^{\frac{1}{\eta}Q(b)}},$$

with $\lambda^\star = \eta \log \sum_a q(a) \exp(Q(a)/\eta)$, where the regular soft-max arises for a uniform distribution $q(a) = \mathcal{U}(a)$ over all arms. That the soft-max policy results from using the KL divergence as a measure of distance has been suggested several times in the literature (Azar and Kappen, 2012; Peters et al., 2010; Still and Precup, 2012). Further increasing $\alpha$ to $\alpha = 2$, we obtain a **linear** policy

$$\pi^\star(a) = q(a)\left(\frac{Q(a) - \lambda^\star + \kappa(a)}{\eta} + 1\right),$$
$$\lambda^\star = \sum_a q(a)(Q(a) + \kappa(a)).$$

As pointed out in Section 3.1, some actions may get zero probability under $\pi^\star$ when $\alpha > 1$. Therefore, in general one needs to keep track of the Lagrange multipliers $\kappa$. However, we can simplify analysis by dropping $\kappa$ for sufficiently large $\eta$. Indeed, if $q(a) > 0$ for all $a \in \mathcal{A}$, there exists a temperature $\eta > \eta_{\min}$ for which $\pi^\star(a) > 0$, and thus the inequality constraints are inactive. This minimum temperature

$$\eta_{\min} = |\min_a A(a)|$$

is defined in terms of the advantage function

$$A(a) \triangleq Q(a) - J(q) = Q(a) - \sum_b q(b)Q(b)$$

which measures agent's gain from taking action $a$ instead of following policy $q$. The baseline $\lambda^\star$ equals the expected return $J(q)$ under policy $q$ when $\kappa = 0$. Thus, for any $\eta > \eta_{\min}$, we obtain a linearly advantage-reweighted policy

$$\pi^\star(a) = q(a) \left( \frac{1 + A(a)}{\eta} \right).$$

For $\alpha \gg 1$, we obtain a new policy, that we call $\epsilon$-**elimination**. Under this policy, all actions get equal probability except for the set of the worst actions that gets probability zero; the size of the set depends on $\alpha$ and $\eta$.

**From $Q_{\max}$ to $Q_{\min}$: effects of $\alpha$ on the baseline.** To complete the picture, we describe the asymptotic behavior of $\lambda^\star$ when the divergence type $\alpha$ is varied. Consider the limit $\alpha \to -\infty$ of large negative $\alpha$. As established in Section 3.1, Lagrange multipliers $\kappa$ vanish for $\alpha < 1$. For sufficiently large $\alpha$, the solution hits the inequality constraint in (6), from which the lower bound on $\lambda^\star$

$$\lambda^\star \geq Q(a) - \eta\epsilon, \qquad \forall a \in \mathcal{A}$$

with $\epsilon = 1/(1-\alpha)$ can be obtained. Given that it holds for all $a$, it also holds for the maximum over $a$. Since the solution lies at the constraint, in the limit one obtains

$$\lim_{\alpha \to -\infty} \lambda^\star = \lim_{\epsilon \to 0} \left\{ \max_a Q(a) - \eta\epsilon \right\} = \max_a Q(a).$$

Table 3: Effects of $\alpha$ on the baseline $\forall \eta > 0$

| Baseline $\lambda$ | $\alpha$ |
|---|---|
| $Q_{\max}$ | $-\infty$ |
| $(\bar{Q}, Q_{\max})$ | $\alpha \leq 1$ |
| log-sum-exp | $\alpha = 1$ |
| $\bar{Q}$ | $\alpha = 2$ |
| $(Q_{\min}, \bar{Q}]$ | $\alpha > 1$ |
| $Q_{\min}$ | $+\infty$ |

Recall from Section 3.2 that for a fixed $\alpha$ the baseline $\lambda^\star$ shifts towards the interior of the feasible set when the temperature $\eta$ is increased, i.e., $\lim_{\eta \to \infty} \lambda^\star = \mathbb{E}_{a \sim q}[Q(a)] \triangleq \bar{Q}$. Therefore, the optimal baseline $\lambda^\star$ for any divergence $\alpha < 1$ lies in the range $\lambda^\star \in (\bar{Q}, Q_{\max})$. The same analysis for $\alpha > 1$ yields $\lambda^\star \in (Q_{\min}, \bar{Q}]$. Table 3 summarizes these asymptotic results.

### 3.3 Empirical evaluation of policy improvement with $f$-divergence in bandits

The effects of the divergence type are demonstrated on a simulated stochastic multi-armed bandit problem. Policy improvement in a single iteration as well as regret over a period of time are compared for different values of $\alpha$.

**Effects of $\alpha$ on policy improvement.** Figure 3 shows the effects of $\alpha$-divergence on policy update. We consider a 10-armed bandit problem with arm values $Q(a) \sim \mathcal{N}(0, 1)$. The temperature is kept fixed at $\eta = 2$ for all values of $\alpha$. Initial policy is uniform and the arm values $Q(a)$ are fixed. Several iterations are shown in the figure for comparison. Extremely large positive and negative values of $\alpha$ correspond to $\epsilon$-elimination and $\epsilon$-greedy policies respectively. Small values of $\alpha$, in contrast, weigh actions according to their values. Policies for $\alpha < 1$ are peaked and heavy-tailed, eventually turning into $\epsilon$-greedy policies when $\alpha \to -\infty$. Policies for $\alpha \geq 1$ are more uniform, but they put zero mass on bad actions, eventually turning into $\epsilon$-elimination policies when $\alpha \to \infty$. For $\alpha \geq 1$, policy iteration may spend a lot of time in the end deciding between two best actions, whereas for $\alpha < 1$ the final convergence is faster.

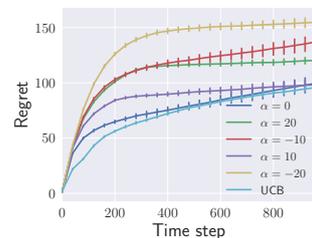

Figure 2: Average regret for various values of $\alpha$.

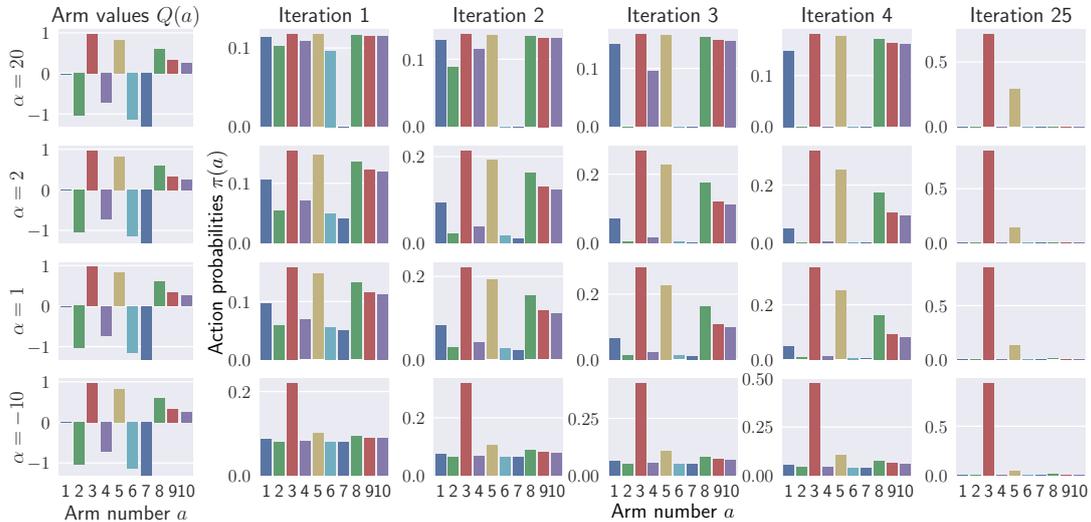

Figure 3: Effects of $\alpha$ on policy improvement. Each row corresponds to a fixed $\alpha$. First four iterations of policy improvement together with iteration 25 are shown in each row. Large positive $\alpha$'s eliminate bad actions one by one, keeping the exploration level equal among the rest. Small $\alpha$'s weigh actions according to their values; actions with low value get zero probability for $\alpha > 1$, but remain possible with small probability for $\alpha \leq 1$. Large negative $\alpha$'s focus on the best action, exploring the remaining ones with equal probability.

**Effects of $\alpha$ on regret.** Average regret $C_n = nQ_{\max} - \mathbb{E}[\sum_{t=0}^{n-1} r_t]$ is shown in Figure 2 for different values of $\alpha$ as a function of the time step $n$ with 95% confidence error bars. We also plot performance of the UCB algorithm (Bubeck and Cesa-Bianchi, 2012) for comparison. The presented results are obtained in a 20-armed bandit environment where rewards have Gaussian distribution $r(a) \sim \mathcal{N}(Q(a), \sigma^2)$ with variance $\sigma^2 = 0.5$. Arm values are estimated from observed rewards and the policy is updated every 20 time steps. The temperature parameter $\eta$ is decreased starting from $\eta = 1$ after every policy update according to the schedule $\eta^+ = \beta\eta$ with $\beta = 0.8$. Results are averaged over 400 runs. In general, extreme $\alpha$'s accumulate more regret. However, they eventually focus on a single action and flatten out. Small $\alpha$'s accumulate less regret, but they may keep exploring sub-optimal actions longer. Values of $\alpha \in [0, 2]$ perform comparably with UCB after around 400 steps, once reliable estimates of values have been obtained.

Figure 4 shows the average regret after a given number of time steps as a function of the divergence type $\alpha$. Best results are achieved for small values of $\alpha$. Large negative $\alpha$'s correspond to $\epsilon$-greedy policies, which oftentimes prematurely converge to a sub-optimal action, failing to discover the optimal action for a long time if the exploration probability $\epsilon$ is small. Large positive $\alpha$'s correspond to $\epsilon$-elimination policies, which may by mistake completely eliminate the best action or spend a lot of time deciding between two options in the end of learning, accumulating more regret. The optimal value of $\alpha$ depends on the time horizon for which one wishes to optimize the policy. The minimum of the curves shifts from slightly negative $\alpha$'s towards the range $\alpha \in [0, 2]$ with increasing time horizon.

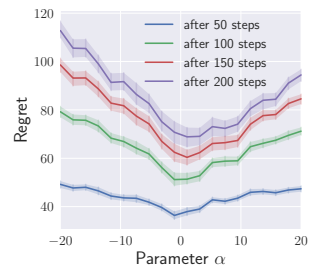

Figure 4: Regret after a fixed time as a function of $\alpha$.

## 4. Policy iteration with state-action $f$-divergence for ergodic MDPs

The methodology described in the previous section can be generalized from the bandit scenario to reinforcement learning. We focus on the average-reward setting (Sutton and Barto, 1998). At every time step, an agent finds itself in a state $s \in \mathcal{S}$ and takes an action $a \in \mathcal{A}$ drawn from a stochastic policy $\pi(a|s)$. Subsequently, the agent transitions from state $s$ to state $s'$ with probability $p(s'|s,a)$ and receives a reward $r(s,a) \in \mathbb{R}$. The fraction of time the agent spends on average in each state $s$ as a result of following policy $\pi$ is given by the steady state distribution $\mu(s)$, which satisfies the stationarity condition

$$\mu(s') = \sum_{s,a} \mu(s)\pi(a|s)p(s'|s,a). \tag{7}$$

The steady state distribution is sometimes denoted by $\mu^\pi(s)$ to emphasize its dependence on the policy. The agent seeks a policy $\pi$ that maximizes the expected return

$$J(\pi) \triangleq \sum_{s,a} \mu(s)\pi(a|s)r(s,a) \tag{8}$$

subject to the constraint (7), provided that both $\pi$ and $\mu$ are probability distributions.

### 4.1 $f$-Divergence constrained policy optimization for ergodic MDPs

We can generalize the bandit problem (4) by adding a single additional constraint here—namely, the stationarity condition (7). This constraint directly yields the value function (Peters et al., 2010), i.e., the policy evaluation step, as the corresponding dual variable. However, for a more straightforward Lagrangian, we rewrite (7) in terms of both policy and state distribution. Assuming we have the state-action distribution $q(s,a) = \mu^{\pi_0}(s)\pi_0(a|s)$ under the old policy $\pi_0(a|s)$ to which we aim to stay similar to with the new distribution $\mu(s)\pi(a|s)$ while achieving a higher expected return, we obtain the following optimization problem

$$\begin{aligned}
\max_{\mu,\pi} \quad & J_\eta(\mu,\pi) = \sum_{s,a} \mu(s)\pi(a|s)r(s,a) - \eta \sum_{s,a} q(s,a)f\left(\frac{\mu(s)\pi(a|s)}{q(s,a)}\right) \\
\text{s.t.} \quad & \sum_{a'} \mu(s')\pi(a'|s') = \sum_{s,a} \mu(s)\pi(a|s)p(s'|s,a), \quad \forall s' \in \mathcal{S}, \\
& \sum_{s,a} \mu(s)\pi(a|s) = 1, \\
& \mu(s)\pi(a|s) \geq 0, \quad \forall s \in \mathcal{S}, \forall a \in \mathcal{A}.
\end{aligned} \tag{9}$$

A similar Lagrangian as for problem (4) can be obtained where the value function $v$ arises as an additional Lagrangian multiplier.

The Lagrangian of problem (9) can be written as

$$L(\mu, \pi, v, \lambda, \kappa) = \sum_{s,a} \mu(s)\pi(a|s)A_v(s,a) - \eta \sum_{s,a} q(s,a) f\left(\frac{\mu(s)\pi(a|s)}{q(s,a)}\right)$$
$$- \lambda \left(\sum_{s,a} \mu(s)\pi(a|s) - 1\right) + \sum_{s,a} \kappa(s,a)\mu(s)\pi(a|s),$$

where $A$ is a function of the dual variables $v$ corresponding to the stationarity constraint (7)

$$A_v(s,a) \triangleq r(s,a) + \sum_{s'} p(s'|s,a)v(s') - v(s). \tag{10}$$

We note that $v$ can be identified with the *value function* and $A$ with the *advantage function*. The solution of Problem (9) can be expressed as a function of the dual variables

$$\mu^\star(s)\pi^\star(a|s) = q(s,a) f'_*\left(\frac{A_v(s,a) - \lambda + \kappa(s,a)}{\eta}\right), \tag{11}$$

which allows extracting the optimal next policy by Bayes' rule

$$\pi^\star(a|s) = \frac{q(s,a) f'_*\left((A_v(s,a) - \lambda + \kappa(s,a))/\eta\right)}{\sum_b q(s,b) f'_*\left((A_v(s,b) - \lambda + \kappa(s,b))/\eta\right)}. \tag{12}$$

The optimal values of the dual variables $\{v, \lambda, \kappa\}$ are found by solving the dual problem

$$\begin{aligned}
\min_{v, \lambda, \kappa} \quad & g(v, \lambda, \kappa) = \eta \sum_{s,a} q(s,a) f^*\left(\frac{A_v(s,a) - \lambda + \kappa(s,a)}{\eta}\right) + \lambda \\
\text{s.t.} \quad & \kappa(s,a) \geq 0, \quad \forall s \in \mathcal{S}, \forall a \in \mathcal{A}, \\
& \arg f^* \in \text{range}_{x \geq 0} f'(x), \quad \forall s \in \mathcal{S}, \forall a \in \mathcal{A}.
\end{aligned} \tag{13}$$

The constraint on $\arg f^*$ is linear because derivative of a convex function is monotone (Boyd and Vandenberghe, 2004). Problem (13) altogether is a convex optimization problem, therefore can be solved efficiently.

A generic policy iteration algorithm employing an $f$-divergence penalty proceeds in two steps. Policy evaluation: the dual problem (13) is solved for a given $q$. Policy improvement: the analytic solution (12) of the primal problem (9) is used to update $\pi$. In the following subsection we consider several particular choices of the divergence function in more detail.

**4.2 Two important special cases: $\alpha = 1$ and $\alpha = 2$**

Similar to the bandit scenario (see Section 3.2), there are two special choices of the $f$-divergence for which we can analytically find the Lagrange multipliers $\lambda$ and $\kappa$ and thus obtain the policy evaluation and policy improvement steps in the simplest form.

**KL divergence ($\alpha = 1$).** With $f = f_1$, policy evaluation (13) turns into

$$\min_v g(v) = \lambda^\star(v) = \eta \log \sum_{s,a} q(s,a) \exp\left(\frac{A_v(s,a)}{\eta}\right) \tag{14}$$

where $g(v) \triangleq L(\mu^\star, \pi^\star, v, \lambda^\star, \kappa^\star)$ is the Lagrangian upon substitution of all variables except for $v$. Remarkably, the baseline $\lambda^\star$ equals the Lagrangian dual function $g$, which is only true in the KL divergence case. The policy improvement step (12) takes the form

$$\pi^\star(a|s) = \frac{q(s,a)\exp\left(A_{v^\star}(s,a)/\eta\right)}{\sum_b q(s,b)\exp\left(A_{v^\star}(s,b)/\eta\right)}. \qquad (15)$$

Note that the baseline $\lambda^\star$ cancels out in the policy but would be included in the state-action distribution.

**Pearson $\chi^2$ divergence ($\alpha = 2$).** When $f = f_2$, one needs to be careful with the dual variables $\kappa$ ensuring the non-negativity of probabilities because they are not necessarily all zero (see Section 3.1). For notational simplicity, we employ *differential advantage* $\tilde{A}_v(s,a)$ defined as a difference between the advantage (10) and the expected return (8)

$$\tilde{A}_v(s,a) = A_v(s,a) - J(\pi_0),$$

following Sutton and Barto (1998). It is straightforward to check that if the temperature $\eta$ is bigger than the absolute value of the minimum differential advantage, $\eta > \eta_{\min} = |\min_{s,a} \tilde{A}_v(s,a)|$, then the non-negativity constraint is satisfied and $\kappa \equiv 0$. The baseline $\lambda^\star$ turns into the average return

$$\lambda^\star = \sum_{s,a} q(s,a) A_v(s,a) = J(\pi_0),$$

becoming independent of $v$, in contrast to the KL divergence case. The policy evaluation step (13) corresponds to minimization of the squared differential advantage (scaled by $2\eta$)

$$\min_v g(v) = \frac{1}{2\eta} \sum_{s,a} q(s,a) \tilde{A}_v^2(s,a) + J(\pi_0), \qquad (16)$$

and the policy improvement step (12) corresponds to the linear in the differential advantage reweighting of the old policy

$$\pi^\star(s,a) = \frac{q(s,a)\left(1 + \tilde{A}_{v^\star}(s,a)/\eta\right)}{\sum_b q(s,b)\left(1 + \tilde{A}_{v^\star}(s,b)/\eta\right)}. \qquad (17)$$

A careful reader shall notice that the solution for the Pearson $\chi^2$-divergence (16, 17) can be obtained as the first-order approximation of the KL divergence solution (14, 15) in the high-temperature limit $\eta \to \infty$. It turns out, an even more general statement holds. Solution (12, 13) with any $f$-divergence turns into the linear-quadratic Pearson $\chi^2$ solution in the limit of high temperatures (small policy update steps). The underlying reason for this is that the Pearson $\chi^2$-divergence is a quadratic approximation of any $f$-divergence around unity. Since pushing $\eta$ towards infinity puts more weight on the divergence penalty in the optimization objective (9), the distance between the old state-action distribution and the new one becomes smaller, justifying the Taylor expansion. Therefore, one should expect to see pronounced differences between various $f$-divergence penalties either for big update steps or when approximations of the value function and the policy are used.

### 4.3 Practical algorithm: sample-based policy iteration with $f$-divergence

Solving problem (13) and computing the optimal next policy (12) would require full knowledge of the environment $\{p(s'|s,a), r(s,a)\}$. In practice, a reinforcement learning agent would need to estimate $p$ and $r$ from experience, which is prohibitively expensive in large state-action spaces. Fortunately, we can rephrase the algorithm entirely in terms of sample averages, thus side-stepping model estimation. We replace expectations by sample averages using the transitions $(s_i, a_i, s'_i, r_i)$ gathered under the current policy. The agent does not need to visit all states—local information under the current policy suffices to compute the averages. Algorithm 1 summarizes the main steps of the $f$-divergence constrained policy iteration.

**Function approximation.** If the dimensionality of the state space is large, one can resort to function approximation methods. We note that relaxing the steady state constraint (7) to a "steady features constraint"

$$\sum_{s'} \mu(s')\phi(s') = \sum_{s,a,s'} \mu(s)\pi(a|s)p(s'|s,a)\phi(s'). \tag{18}$$

naturally introduces a linear function approximation $v_\theta(s) = \theta^T \phi(s)$ with fixed features $\phi \colon \mathcal{S} \to \mathbb{R}^m$ while the parameters $\theta$ are the new Lagrangian multipliers (Peters et al., 2010). To our knowledge, such a constraint is the only way to introduce function approximation in the same way into the primal optimization problem (9), the next policy (12), and the dual optimization problem (13).

**Implementation details.** Several practical improvements make Algorithm 1 faster and robuster. First, the temperature parameter $\eta$ should be decayed with iterations in order to guarantee convergence to a solution of the original non-penalized problem. In our experiments, simple exponential decay was sufficient; however, more complex adaptive schemes can be devised. Second, for $\alpha$-divergences with $\alpha \leq 1$, one can omit the dual variables $\kappa(s,a)$ because they are all equal to zero (see Section 3.1). Moreover, one can omit $\kappa(s,a)$ even for $\alpha > 1$ if $\eta$ is sufficiently big. The exact value of $\eta_{\min}$ above which $\kappa$ can be ignored depends on the scale of the rewards; in the special case $\alpha = 2$, there is an explicit condition on $\eta_{\min}$ (see Section 4.2). Third, the estimate of the advantage $A_{v^\star}(s,a)$ for actions that were not sampled under the current policy can be set to the baseline $\lambda^\star$, which serves as a proxy for the expected reward; in our experiments this strategy performed best. Fourth, introducing a little slack in the linear constraint on the domain of the convex conjugate function improves robustness and helps the optimizer.

### 4.4 Empirical results on ergodic MDPs

We evaluate our policy iteration algorithm with $f$-divergence on standard grid-world reinforcement learning problems from OpenAI Gym (Brockman et al., 2016). The environments that terminate or have absorbing states are restarted during data collection in order to ensure ergodicity. Figure 5 demonstrates learning dynamics on different environments for various choices of the divergence function. Parameter settings and other implementation details can be found in Appendix A. In summary, one can either promote risk averse behavior by choosing $\alpha < 0$, which may, however, result in suboptimal exploration, or one can

---

**Algorithm 1:** Policy iteration with any type of $f$-divergence
   **Input**: Initial policy $\pi_0(a|s)$, divergence function $f$, temperature $\eta$
   **define** Dual Function $\hat{g}(v, \lambda, \kappa)$:
      **for** *every sample* $i \leftarrow 1$ to $N$ **do**
         $\hat{A}(s_i, a_i) \leftarrow \hat{A}(s_i, a_i) + (r_i + v(s'_i) - v(s_i))$
         $n(s_i, a_i) \leftarrow n(s_i, a_i) + 1$
      **for** *every state-action pair* $(s, a) \in \mathcal{D}$ **do**
         $\hat{A}(s, a) \leftarrow \hat{A}(s, a)/n(s, a)$
         $\hat{y}(s, a) \leftarrow (\hat{A}(s, a) - \lambda + \kappa(s, a))/\eta$
      **return** $\frac{\eta}{N} \sum_{i=1}^{N} f^*(\hat{y}(s_i, a_i)) + \lambda$
   **foreach** *policy update k* **do**
      **Sampling**: Gather $\mathcal{D} = \{(s_i, a_i, s'_i, r_i), i = \overline{1, N}\}$ under current policy $\pi_k(a|s)$
      **Policy Evaluation**: Minimize dual function $\hat{g}$ defined above
         $\{v^\star, \lambda^\star, \kappa^\star\} \leftarrow \arg\min_{v, \lambda, \kappa} \hat{g}(v, \lambda, \kappa)$ s.t. $\kappa(s, a) \geq 0$, $\hat{y}(s, a) \in \text{range}_{x \geq 0} f'(x)$
      **Policy Improvement**: **for** *every state-action pair* $(s, a) \in \mathcal{D}$ **do**
         $\pi_{k+1}(a|s) \leftarrow \frac{\pi_k(a|s) f'_*(\hat{y}(s, a; v^\star, \lambda^\star, \kappa^\star))}{\sum_b \pi_k(b|s) f'_*(\hat{y}(s, b; v^\star, \lambda^\star, \kappa^\star))}$
   **Output**: Optimal policy $\pi^\star(a|s)$

---

promote risk seeking behavior with $\alpha > 1$, which may lead to overly aggressive elimination of options. Our experiments suggest that the optimal balance should be found in the range $\alpha \in [0, 1]$. It should be pointed out that the effect of the $\alpha$-divergence on policy iteration is not linear and not symmetric with respect to $\alpha = 0.5$, contrary to what one could have expected given the symmetry of the $\alpha$-divergence as a function of $\alpha$ pointed out in Section 2. That is, switching from $\alpha = -3$ to $\alpha = -2$ may have little effect on policy iteration, whereas switching from $\alpha = 3$ to $\alpha = 4$ may have a much more pronounced influence on the learning dynamics.

### 4.5 Relation to the mean squared Bellman error minimization

We have seen in Section 4.2 that policy evaluation with the Pearson $\chi^2$-divergence penalty, corresponding to the dual optimization problem (16), is equivalent to minimization of the mean squared differential advantage (MSDA). Here, we establish the relation between the MSDA objective and the classical mean squared Bellman error (MSBE) minimization (Sutton and Barto, 1998).

In order to state precisely how the MSDA and MSBE objectives are related, we need to recall several definitions. Assume that the value function $v \colon \mathcal{S} \to \mathbb{R}$ is parameterized by a vector $\theta$ as discussed in Section 4.3, and let an agent follow a policy $\pi_0$ which yields a state-action distribution $q(s, a) = \mu^{\pi_0}(s)\pi_0(a|s)$. In discounted infinite horizon setting, one defines the *temporal difference (TD) error* as $\delta_\theta(s, a, s') = r(s, a) + \gamma v_\theta(s') - v_\theta(s)$ where $\gamma \in (0, 1)$ is a discount factor. In the average-reward setting, this TD error is typically replaced by the *differential TD error*

$$\tilde{\delta}_\theta(s, a, s') = r(s, a) - \bar{r} + v_\theta(s') - v_\theta(s)$$

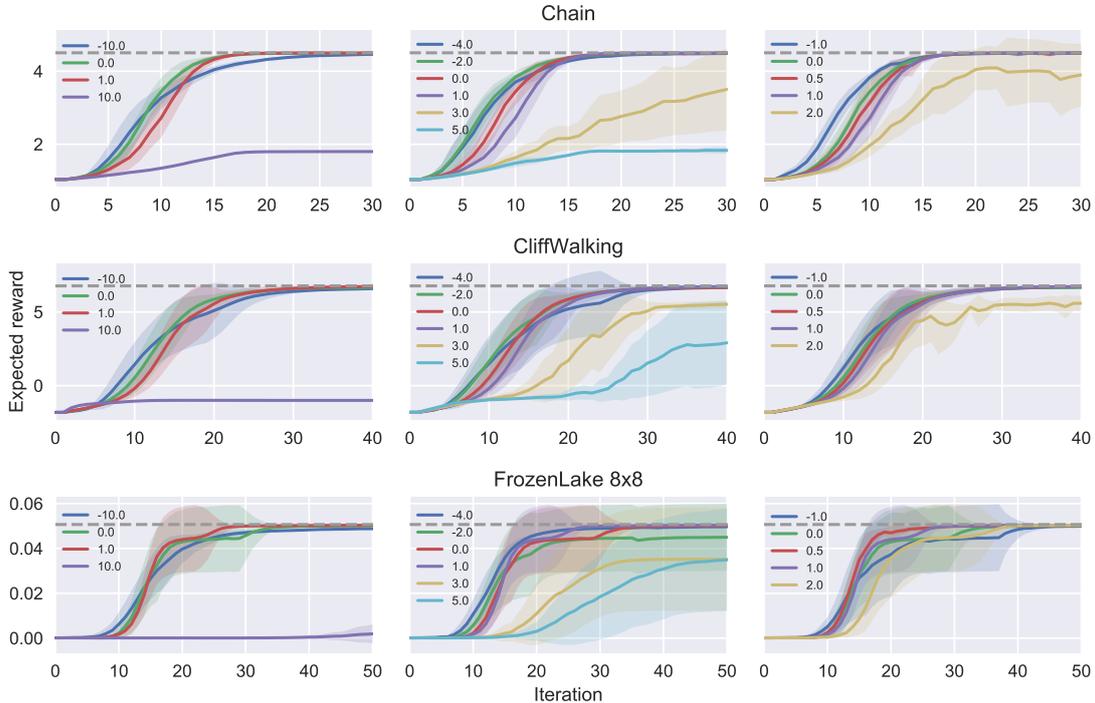

Figure 5: Effects of $\alpha$-divergence on policy iteration. Each row corresponds to a given environment. Results for different values of $\alpha$ are split into three subplots within each row, from the more extreme $\alpha$'s on the left to the more refined values on the right. In all cases, more negative values $\alpha < 0$ initially show faster improvement because they immediately jump to the mode and keep the exploration level low; however, after a certain umber of iterations they get overtaken by moderate values $\alpha \in [0, 1]$ that weigh advantage estimates more evenly. Positive $\alpha > 1$ demonstrate high variance in the learning dynamics because they clamp the probability of good actions to zero if the advantage estimates are overly pessimistic, never being able to recover from such a mistake. Large positive $\alpha$'s may even fail to reach the optimum altogether, as exemplified by $\alpha = 10$ in the plots. The most stable and reliable $\alpha$-divergences lie between the reverse KL ($\alpha = 0$) and the KL ($\alpha = 1$), with the Hellinger distance ($\alpha = 0.5$) outperforming both of them on the FrozenLake environment.

that differs from the "usual" TD error in two ways: first, there is no discount factor, and second, the expected reward under the current policy

$$\bar{r} = \sum_{s,a} q(s,a) r(s,a) = \sum_{s,a} \mu^{\pi_0}(s) \pi_0(a|s) r(s,a) = J(\pi_0)$$

is subtracted in order to ensure that the mean of the differential TD error is zero

$$\sum_{s,a} \mu^{\pi_0}(s) \pi_0(a|s) p(s'|s,a) \tilde{\delta}_\theta(s,a,s') = 0.$$

In the following, we only consider differential versions of the errors. By averaging over the next states $s'$ in the differential TD error, one obtains the *differential advantage*

$$\tilde{A}_\theta(s, a) = \sum_{s'} p(s'|s, a) \tilde{\delta}_\theta(s, a, s').$$

By integrating out the action $a$ in the differential advantage, one obtains the *differential Bellman error*

$$\tilde{\varepsilon}_\theta(s) = \sum_a \pi_0(a|s) \tilde{A}_\theta(s, a).$$

In total, there are three different error signals: differential TD error (DTDE), differential advantage (DA), and the differential Bellman error (DBE). Each of them can be used for learning. Creating a *mean squared* (MS) objective out of each of them, we obtain three objective functions:

$$\text{MSDTDE}(\theta) = \sum_{s,a,s'} \mu^{\pi_0}(s) \pi_0(a|s) p(s'|s, a) \left[ \tilde{\delta}_\theta(s, a, s') \right]^2,$$

$$\text{MSDA}(\theta) = \sum_{s,a} \mu^{\pi_0}(s) \pi_0(a|s) \left[ \sum_{s'} p(s'|s, a) \tilde{\delta}_\theta(s, a, s') \right]^2,$$

$$\text{MSDBE}(\theta) = \sum_s \mu^{\pi_0}(s) \left[ \sum_{a,s'} \pi_0(a|s) p(s'|s, a) \tilde{\delta}_\theta(s, a, s') \right]^2.$$

Thus, the novel MSDA objective fits in-between the TD error and the Bellman error minimization. Performing stochastic gradient descent (SGD) on these objectives leads to various flavors of the residual gradient (RG) algorithm (Baird, 1995), recently surveyed by Dann et al. (2014). Curiously, all three objectives result in the naive residual gradient algorithm (Sutton and Barto, 1998) when expectations are "naively" replaced by one-sample estimates.

We realize that the dual for $\alpha = 2$, i.e., (16) is exactly the MSDA objective. Interestingly, since the dual objective (13) tends towards the MSDA objective in the limit of high temperatures $\eta \to \infty$ for any choice of the divergence function $f$, dual minimization can be viewed as a continuous generalization of the MSDA minimization to finite temperatures; therefore, SGD on the dual objective (13) is a continuous generalization of the RG algorithm to finite temperatures.

## 5. Conclusion

In this paper, we have developed a framework for deriving new and established policy improvement and policy iteration algorithms by constraining policy change using the $f$-divergence. First, we described the key idea in the bandit scenario. Looking at the smooth curve of $\alpha$-divergences in the space of $f$-divergences, we were able to see the connection between many classical action selection strategies as well as discover several new ones. The optimal choice of $\alpha$ depends on the time horizon, since different $\alpha$'s corresponds to different exploration strategies. The soft-max policy update rule, employed by the Exp3 algorithm (Bubeck and Cesa-Bianchi, 2012) and the gradient bandit algorithm (Sutton and Barto, 1998), is a special case of the $f$-divergence constrained policy improvement corresponding to the KL divergence.

After the bandit case, we considered the average-reward reinforcement learning problem, for which we found both (i) an efficient sample-based solution and (ii) the important new insight that the choice of the $f$-divergence does not only imply the type of policy improvement step but also the compatible policy evaluation step. In particular, we have shown that employing the soft-max policy for the actor implies that the log-sum-exp function is used for policy evaluation by the critic (somewhat similar to REPS (Peters et al., 2010)). Similarly, in our framework, policy evaluation by minimizing the mean squared differential advantage, which is closely related to the mean squared Bellman error, results from the choice of the $\alpha = 2$-divergence and comes along with a new, compatible policy update rule. Thus, it is likely that this approach will still yield a larger number of new policy improvements along with their compatible critics derived by just choosing another $f$-divergence.

## Acknowledgments

This project has received funding from the European Union's Horizon 2020 research and innovation programme under grant agreement No 640554.

**Appendix A.**

In all experiments, the temperature parameter $\eta$ is exponentially decayed $\eta_{i+1} = \eta_0 a^i$ in each iteration $i = 0, 1, \ldots$. The choice of $\eta_0$ and $a$ depends on the scale of the rewards and the number of samples collected per policy update. Tables for each environment list these parameters along with the number of samples per policy update, the number of policy iteration steps, and the number of runs for averaging the results. Where applicable, environment-specific settings are also listed.

| Parameter | Value |
| --- | --- |
| Number of states | 8 |
| Action success probability | 0.9 |
| Small and large rewards | (2.0, 10.0) |
| Number of runs | 10 |
| Number of iterations | 30 |
| Number of samples | 800 |
| Temperature parameters $(\eta_0, a)$ | (15.0, 0.9) |

Table 4: Chain environment.

| Parameter | Value |
| --- | --- |
| Punishment for falling from the cliff | -10.0 |
| Reward for reaching the goal | 100 |
| Number of runs | 10 |
| Number of iterations | 40 |
| Number of samples | 1500 |
| Temperature parameters $(\eta_0, a)$ | (50.0, 0.9) |

Table 5: CliffWalking environment.

| Parameter | Value |
| --- | --- |
| Action success probability | 0.8 |
| Number of runs | 10 |
| Number of iterations | 50 |
| Number of samples | 2000 |
| Temperature parameters $(\eta_0, a)$ | (1.0, 0.8) |

Table 6: FrozenLake environment.